\pdfoutput=1

\documentclass[11pt]{article}

\usepackage[final]{acl}

\usepackage{times}
\usepackage{latexsym}

\usepackage[T1]{fontenc}

\usepackage[utf8]{inputenc}

\usepackage{microtype}

\usepackage{kotex}
\usepackage{booktabs}
\usepackage{graphicx}
\usepackage{amssymb}
\usepackage{amsmath}
\usepackage{multirow}
\usepackage{comment}
\usepackage{adjustbox}
\usepackage{mathtools}
\usepackage[normalem]{ulem}
\usepackage{pifont}
\usepackage{xcolor}
\usepackage{color, colortbl}
\usepackage[nomessages]{fp}

\newcommand{\ie}{{\it i.e.}}%
\newcommand{\eg}{{\it e.g.}}%

\newcommand\Tstrut{\rule{0pt}{2.2ex}}       
\newcommand\Bstrut{\rule[-0.6ex]{0pt}{0pt}} 
\newcommand{\TBstrut}{\Tstrut\Bstrut} 
\newcommand\DTstrut{\rule{0pt}{5.2ex}}       
\newcommand\DBstrut{\rule[-1.6ex]{0pt}{0pt}} 
\newcommand{\DTBstrut}{\DTstrut\DBstrut} 

\definecolor{Gray065}{gray}{0.65}
\definecolor{Gray07}{gray}{0.7}
\definecolor{Gray075}{gray}{0.75}
\definecolor{Gray08}{gray}{0.8}
\definecolor{Gray085}{gray}{0.85}
\definecolor{Gray09}{gray}{0.9}
\definecolor{Gray095}{gray}{0.95}

\newcommand{\maxnum}{100.00}
\newlength{\maxlen}
\newcommand{\databar}[2][blue!30]{%
  \settowidth{\maxlen}{\maxnum}%
  \addtolength{\maxlen}{\tabcolsep}%
  \FPeval\result{round(#2/\maxnum:4)}%
  \rlap{\color{blue!30}\hspace*{-.5\tabcolsep}\rule[-.05\ht\strutbox]{\result\maxlen}{.95\ht\strutbox}}%
  \makebox[\dimexpr\maxlen-\tabcolsep][r]{#2}%
}

\title{Towards Zero-Shot Functional Compositionality of Language Models}

\author{
\quad Hangyeol Yu$^{1}$\thanks{\quad Equal Contribution.}
\quad Myeongho Jeong$^{1}$\footnotemark[1]
\quad Jamin Shin$^{2}$\thanks{\quad Corresponding Authors. Most work was done while all corresponding authors were at Riiid AI Research.}\\
\quad \textbf{Hyeongdon Moon}$^{1}$
\quad \textbf{Juneyoung Park}$^{1}$
\quad \textbf{Seungtaek Choi}$^{1}$\footnotemark[2]\\ 
$^1$Riiid AI Research\\
$^2$NAVER AI Lab \\
\texttt{\{hangyeol.yu,myeongho.jeong\}@riiid.co},\\
\texttt{jayshin.nlp@gmail.com},
\texttt{seungtaek.choi@riiid.co}
}
\begin{document}
\maketitle

\vspace{-30mm}
\begin{abstract}
Large Pre-trained Language Models (PLM) have become the most desirable starting point in the field of NLP, as they have become remarkably good at solving many individual tasks.
Despite such success, in this paper, we argue that current paradigms of working with PLMs are neglecting a critical aspect of modeling human intelligence: \textbf{functional compositionality}. Functional compositionality -- the ability to compose learned tasks -- has been a long-standing challenge in the field of AI (and many other fields) as it is considered one of the hallmarks of human intelligence.
An illustrative example of such is cross-lingual summarization, where a bilingual person (English-French) could directly summarize an English document into French sentences \textit{without} having to translate the English document or summary into French explicitly.
We discuss why this matter is an important open problem that requires further attention from the field. 
Then, we show that current PLMs (\eg, GPT-2 and T5) don't have functional compositionality yet and it is far from human-level generalizability.
Finally, we suggest several research directions that could push the field towards \textit{zero-shot} functional compositionality of language models.
\footnote{Our code is released here: \url{https://github.com/jshin49/tclm}.}
\end{abstract}
\section{Introduction}
\begin{figure}[!t]
  \centering
  \includegraphics[clip,width=\columnwidth]{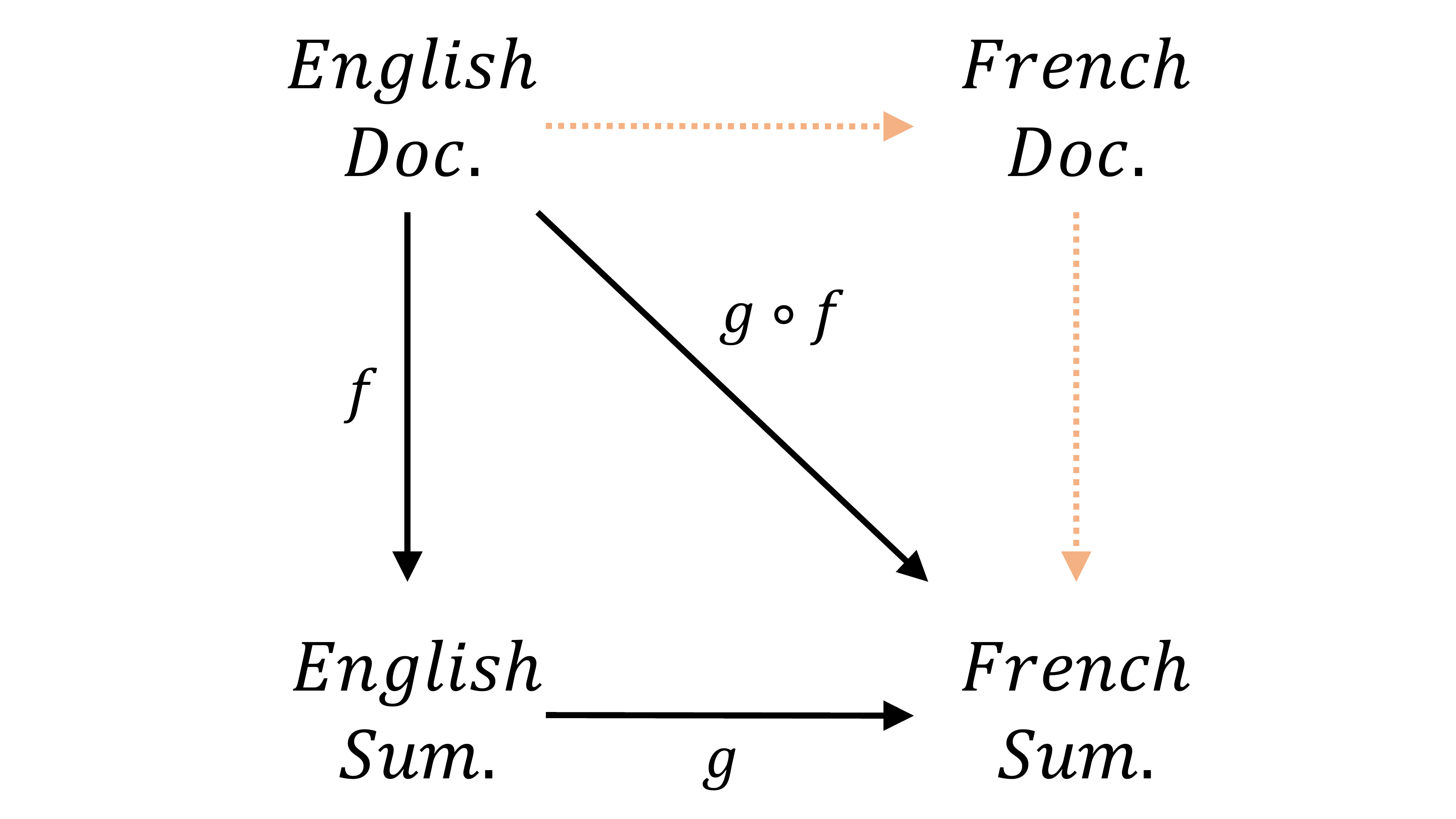}
  \caption{A simple representation of Cross-lingual Summarization as a function composition ($g \circ f$) of summarization ($f$) and translation ($g$). Sequentially conducting summarization and then translation corresponds to the traditional pipeline architecture, while a model with functional compositionality should directly follow the diagonal edge. Dashed edges are not covered in this work.}
  \label{fig:xls_composition}
\end{figure}

Recently developed large Pre-trained Language Models (PLM)~\cite{devlin2019bert,brown2020language,raffel2020exploring} or Foundation Models~\cite{bommasani2021opportunities} have not only achieved state-of-the-art performance through transfer learning in various benchmarks like GLUE~\cite{wang2018glue} and SuperGLUE~\cite{wang2019superglue} but have also shown dramatic improvements in few-shot and zero-shot learning~\cite{alex2021raft,liu2022few}.

It is clear that we have come a long way, but we are still far from achieving human-level generalizability.
Compositionality, one of the skills for achieving human-level generalizability, has been explored in many works. \cite{nayak2022learning} achieves compositionality between various attributes and objects in visual recognition. Also, \cite{logeswaran2021learning} studies compositional generalization in reinforcement learning by composing language instructions via attention. However, such works are not enough to achieve human-level generalizability.
We argue that there has not been enough focus on how humans naturally compose tasks or functions that they learned~\cite{singh1991transfer,li2020solving}.
In this position paper, inspired by composite functions from mathematics, we introduce a perspective called \textit{functional compositionality}.
This is a different concept from the traditional discussions about the semantic compositionality of human language, where atomic meanings are composed to create new semantics~\cite{liang2013lambda,pasupat2015compositional,kim2020cogs}\footnote{We will cover this more in Section~\ref{sec:related}.}.
Instead, our scope of functional compositionality refers to the end-to-end chaining of two different text-to-text transformations, just like function composition from mathematics.
As many NLP tasks can be reformulated as text-to-text tasks~\cite{raffel2020exploring,brown2020language,alex2021raft}, we believe this is not a small scope.

The most illustrative example is Cross-Lingual Summarization (XLS)~\cite{wang2022survey}. 
As shown in Figure~\ref{fig:xls_composition}, bilingual people should naturally be able to compose their skills of summarization and translation in order to summarize an English document into a French sentence, \textit{without requiring specialized training} to do so. 
What we expect from large versatile PLMs is also similar.
A model that can summarize English documents and translate English to French should be able to create French summary sentences or even summarize French documents \textit{without explicit supervision} of such tasks\footnote{We use the terms function and task interchangeably.}.

However, as we will show later, this is not possible yet in an \textit{end-to-end} fashion. As an alternative, we explore teaching PLMs how to compose tasks. Our fundamental assumption is that the knowledge of how to compose atomic tasks chosen within a restricted set can be transferred to unseen combinations of tasks. This gives a potential direction toward human-level generalizability, but there is still a long way to go.

In this work, we attempt to answer the question of \textit{how far current text-to-text PLMs are from achieving zero-shot functional compositionality}. Our findings can be summarized as such:
\begin{itemize}
    \setlength\itemsep{-0.3em}
    \item Current PLMs have difficulty in composing text-to-text functions end-to-end by zero-shot.
    \item However, they were able to ``Learn to Compose (L2C)'' when explicitly trained to do so on StylePTB~\cite{lyu2021styleptb}.
    \item The L2C method also showed potential to work well with recent parameter-efficient fine-tuning methods, but struggled in transferring the learned task-composing skills to other more difficult benchmarks like WikiLingua~\cite{ladhak2020wikilingua}.
\end{itemize}

Through this work, we shed light on a new research direction for large PLMs in order to advance toward human-level generalizability.
\section{Background and Related Work}
\label{sec:related}

Compositionality has been a long-standing challenge in AI and has been well-studied in other many fields, such as theory of computation, linguistics, philosophy, and mathematics.
In this section, we first cover existing works on semantic compositions (or compositional semantics). Then, we introduce the concept of functional compositions and its distinction from semantic compositions. Finally, we discuss its importance and close this section clarifying the scope of functions considered in this paper.

\subsection{Semantic Compositions}
The principle of compositionality~\cite{pelletier1994principle} has been widely studied in many fields, 
In compositional semantics~\cite{janssen1997compositionality}, the meanings of words or phrases are determined by \textit{combining} the meanings of their sub-words or sub-phrases, and this principle usually holds when syntactic factors play in the increased complexity of a sentence~\cite{szabo2004compositionality}.
As such, this field has often been studied in semantic parsing where complex syntactic rules play a major role in natural language understanding~\cite{liang2013lambda,pasupat2015compositional,yin2021compositional,gupta2018semantic,oren2020improving,kim2020cogs,szpektor2020dynamic,parthasarathi2020task}.
Meanwhile, there was no consensus on whether neural networks are able to generalize compositionally.
Hence, \citet{hupkes2020compositionality} discusses this subject in depth by presenting a set of definitions and tests that is grounded on a vast amount of linguistic and philosophical literature, using probabilistic context-free grammar datasets.
Another very good example can also be found in visual recognition~\cite{misra2017red,wang2019task,naeem2021learning,purushwalkam2019task,logeswaran2021learning,cohen2021learning,nayak2022learning}.
Here, if a model understands the meaning of the phrases ``grey elephant'' and ``blue bottle'', they test if it also generalizes to new vision-language concepts like ``blue elephant''.

\subsection{Functional Compositions}
Inspired by \textit{closed-form composite functions} from mathematics, we define a functional composition as the end-to-end chaining of any two tasks. 
Figure~\ref{fig:xls_composition} illustrates this concept very well. Instead of taking two side edges (like a pipeline) to conduct cross-lingual summarization, a model with funtional compositionality should take the diagonal edge.
Just like a closed-form composite function, we should be able to compute only once while the output is the same as or better than sequentially applying all functions.

This problem has been somewhat discussed in various kinds of literature. 
Task decomposition has been a big problem in reinforcement learning literature~\cite{sahni2017learning,devin2019plan,li2020solving,lee2018composing,mendez2021modular}.
Zero-shot cross-lingual transfer is directly related to our definition of functional composition even though there was no in-depth discussion on it~\cite{conneau2019cross,conneau2020unsupervised,zhao2021discrete,ansell2021composable,barbieri2021xlm,wu2022learning,gritta2022crossaligner}. 
Recently, a compositional style transfer dataset has been released~\cite{lyu2021styleptb}. 
Finally, aggregation of entire network parameters~\cite{madotto2020attention,choshen2022fusing} and adaptive integration of task-specific parameters~\cite{pfeiffer2021adapterfusion,zhang2022continual} can also be viewed as an instance of functional compositions.

\subsection{Why functional compositionality?}
The most obvious benefits of functional compositions would be more efficient inference during deployment than pipelines.
More importantly, if a model can (functional) compositionally generalize, this means that collecting expensive datasets like WikiLingua~\cite{ladhak2020wikilingua} for XLS may no longer be necessary.
Ideally, we can train a model only on the more abundant datasets of the decomposed tasks.

We believe the impact of this matter is very timely as our definition is not just limited to text sequences.
The demand for multi-modal language models has been rapidly increasing in both the industry and research community, and there have already been many successful cases in various tasks: Dall-E 2~\cite{dalle2} and StableDiffusion~\cite{stable_diffusion} for realistic text-to-image synthesis, and Make-A-Video~\cite{singer2022make} for text-to-video synthesis.
However, such models often require a significantly large amount of multi-modal paired data (and model size) that often drastically exceeds academic budgets.
Therefore, expanding these models to languages other than English would require a tremendous amount of data and model parameters.
Furthermore, many multi-modal tasks that were solved through pipelines have recently been tackled with end-to-end models, such as Machine Translation directly on images~\cite{google_image_translation} or on Speech~\cite{google_speech_translation} from Google.
We believe creating models that generalize well to functional compositions will allow what is mentioned at a much lower cost.

\subsection{Scope of Function}
In this paper, we narrow down the scope of function to a text-to-text function with no side effects; the input is a text and so is the output.
Recent works~\cite{raffel2020exploring, brown2020language, sanh2021multitask} build unified learning frameworks by casting various NLP functions as text-to-text functions.
This would include most of the well-known text generation tasks like machine translation, text summarization, style transfer, conversation, etc.
These text-to-text functions allow us using a consistent training objective for various NLP functions.
As a future direction, we can also trivially extend this definition to \textit{any sequence-to-sequence tasks} like Automatic Speech Recognition or text-to-image tasks or even Image Captioning -- as we can consider an image as a sequence of patches~\cite{dosovitskiy2020image}.

\section{Methods}
\label{sec:method}
\begin{figure*}[!t]
  \centering
  \includegraphics[clip,width=0.95\textwidth]{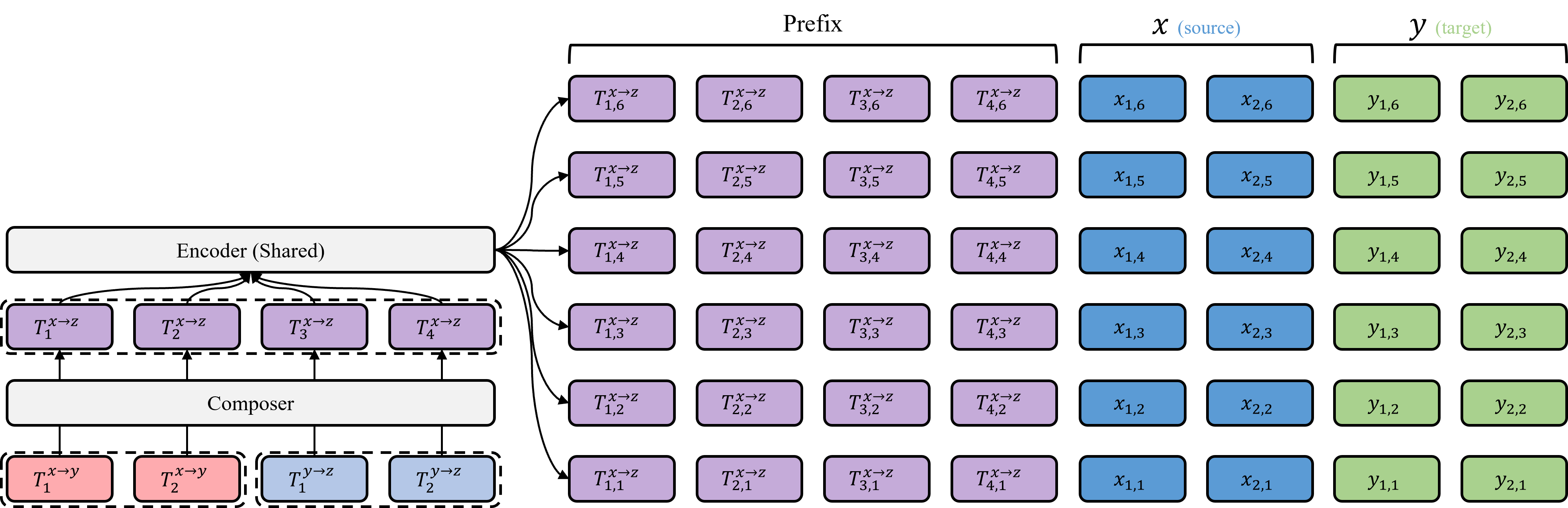}
  \caption{The overall architecture of prefix composition for composite tasks. It is a non-destructive composition with task-specific parameters, by using a learnable self-attention layer, which is the part marked as composer in the figure. Note that unlike the original prefix tuning, a single MLP encoder is shared for multiple atomic tasks}
  \label{fig:architecture}
\end{figure*}
We suggest three ways to let a PLM compose an unseen (zero-shot) combination of tasks while each task is already learned. Thus, every method assumes that the model is trained on atomic tasks via multi-task learning. Note that due to their recent successes, we mainly conduct our experiments on prompt-based language models such as T5~\cite{raffel2020exploring} and GPT~\cite{brown2020language}. Especially, they show strong performance with zero-shot and few-shot learning on multi-task benchmarks. Therefore, our three methods are mostly based on recent prompt-based learning. This section describes how each method works in detail.

\subsection{Prompt-based Fine-tuning (\textsc{Prompt})}

Prompt-based fine-tuning (\textsc{Prompt})~\cite{lester2021power,han2021ptr} has recently been the most popular way to fine-tune a PLM. To specify which task the model should perform, a task-specific (text) prefix is added to the original input sequence before feeding it to the model.
Suppose we have a language model parameterized by $\phi$. Normally, prompting is simply prepending a fixed series of tokens, $Z$, to the input $X$. We shall denote the concatenated sequence as $[Z;X]$. Then, the model tries to maximize the likelihood of the correct $Y$, \ie, $Pr_{\phi}(Y|[Z;X])$.
In this setting, all of the model parameters are completely shared learning multiple downstream tasks.
\paragraph{Prompt Composition}
Suppose we have a PLM that is trained with multiple atomic tasks by \textsc{Prompt} method. In other words, the model is already adapted to a set of prompts where each prompt corresponds to one of the tasks. To let the model compose those tasks, a very natural attempt is to mix learned prompts semantically. To be specific, we automatically generate such prompts with template-based concatenations\footnote{We also explore manual writing of the prompts, like ``\texttt{remove all prepositional phrases and change to future tense}'' for style transfer and ``\texttt{summarize into French:}'' for cross-lingual summarization. However, we empirically found that the template-based concatenations outperformed the manual writings. We posit that such counter-intuitive behavior stems from the large diversity of natural language instructions, making it harder to focus on learning how to compose the tasks.}, such as ``\texttt{\{prompt1\} then \{prompt2\}:}'',  ``\texttt{\{prompt2\} after \{prompt1\}}'', or ``\texttt{\{prompt1\}}+\texttt{\{prompt2\}}''.

\subsection{Prefix Tuning (\textsc{Prefix})}
One question that naturally comes up with the idea of \textsc{Prompt} is whether we can learn compositionality in conjunction with recent parameter-efficient fine-tuning methods~\cite{pfeiffer2021adapterfusion,liu2022few} of large language models. Prefix tuning~\cite{li2021prefix} is one of those successful methods. To learn a specific atomic task $t$, it keeps language model parameters $\phi$ frozen, but tunes a small continuous task-specific vector $P_t$ (called prefix) and a multi-layer perceptron $\text{MLP}_{\theta_{t}}$ parameterized by $\theta_t$. Then, the hidden representation $h_i$ of $i$-th token at each layer is computed as follows:
\begin{align}
    h_i = 
        \begin{cases}
            \text{MLP}_{\theta_{t}}(P_t[i, :]), &\text{if } i \in \texttt{idx}_{t}, \\
            \text{LM}_{\phi}(z_i, h_{<i}), &\text{otherwise,}
        \end{cases}
\end{align}
where $\texttt{idx}_t$ denotes the indices of prefix vectors in the given sequence, and $z_i$ is $i$-th input token.
Further details can be found in~\cite{li2021prefix}. 

\paragraph{Prefix Composition}
Inspired by AdapterFusion~\cite{pfeiffer2021adapterfusion}, we explore non-destructive compositions with task-specific parameters, by using a self-attention layer. Specifically, suppose there are two atomic tasks, $t_1$ and $t_2$, and corresponding prefix vectors, $P_{t_1}$ and $P_{t_2}$. Let $t_1 + t_2$ denote the new target task, which is a functional composition of $t_1$ and $t_2$. To get a new prefix vectors, we use self-attention~\cite{vaswani2017attention} as illustrated in Figure~\ref{fig:architecture}, \eg, $P_{t_1 + t_2} = \text{Attn}_\eta([P_{t_1};P_{t_2}])$ where $\eta$ denotes set of additional learnable self-attention parameters. Note that since $\eta$ is randomly initialized, this type of composition cannot be done without training.

One modification from the original implementation of prefix tuning is that we share a single MLP encoder for multiple atomic tasks, \ie, ($\theta_{t_1} = \theta_{t_2} = \theta)$.
Intuitively, it can be thought of as separating the roles of previous prefix tuning into learning how to perform a task (by $P_t$) and how to distribute the task vector to different transformer layers (by $\text{MLP}_{\theta}$).

\subsection{Pipeline (\textsc{Pipeline})}
\textsc{Pipeline}, the method of serving two different models sequentially following a certain order, is a straightforward implementation of function composition. As this approach requires no extra learning cost to compose various tasks, it has been preferred as a strong baseline, \eg, \textsc{translate-test} in XNLI~\cite{conneau2018xnli}.
Nevertheless, it has clear limitations: 1) calling multiple models in a sequence is computationally expensive, 2) the errors can be accumulated between the sub-tasks, and 3) further training on the target composite task cannot be performed in an end-to-end manner.

Furthermore, it is noteworthy that this method is sensitive to the order of sub-tasks. 
For instance, from StylePTB data (Table~\ref{tab:styleptb}), consider composing PPR (removing prepositional phrases) and PTA (voice switch from passive to active), and applying it to a sentence 
``1,214 cars were sold last year by luxury automakers in the U.S.''. 
Then, the pipeline (PPR $\rightarrow$ PTA) first erases the prepositional phrase ``by luxury automakers in the U.S.'' before voice change. 
The remaining sentence itself cannot be rewritten in active voice since the deleted part includes the subject in the final resulting sentence.
On the other hand, the other pipeline of reverse order (PTA $\rightarrow$ PPR) can easily lead to the proper sentence ``Luxury automakers sold 1,214 cars last year.''.
Another example is from XLS. In general, we can summarize-then-translate, but cannot translate-then-summarize (Figure~\ref{fig:xls_composition}), as document-level translation is very challenging. 
We will further explore such order sensitivity of \textsc{Pipeline} in the later discussion (Section~\ref{subsec:lm_can_zeroshot}).

\section{Experiment Setting}
\begin{table*}[t!]
\begin{center}
\resizebox{0.8\textwidth}{!}{
\begin{tabular}{l l c c c}
  \hline
  \noalign{\hrule height0.8pt} 
  Category & Change & Abbreviation & Description  & \# of samples (train/valid/test) \TBstrut \\
  \hline\noalign{\hrule height0.8pt} 
  \multirow{7}{*}{\textbf{Syntax}}
  & \multirow{3}{*}{Tense}
  & TFU & To future tense & 9279 / 1013 / 1006  \TBstrut \\ 
  & & TPR & To present tense & 5564 / 645 / 643  \TBstrut \\
  & & TPA & To past tense & 4684 / 511 / 502  \TBstrut \\
  \cline{2-5}
  & \multirow{2}{*}{Voice}
  & ATP & Active to passive & 2533 / 278 / 284  \TBstrut \\ 
  & & PTA & Passive to Active & 2533 / 278 / 284  \TBstrut \\
  \cline{2-5}
  & \multirow{2}{*}{PP Front Back}
  & PFB & PP front to back & 426 / 23 / 26  \TBstrut \\ 
  & & PBF & PP back to front & 426 / 23 / 27  \TBstrut \\
  \hline\noalign{\hrule height0.8pt} 
  \multirow{2}{*}{\textbf{Semantic}}
  & \multirow{1}{*}{ADJ/ADV Removal}
  & ARR & ADJ or ADV Removal & 4639 / 273 / 276  \TBstrut \\ 
    \cline{2-5}
  & \multirow{1}{*}{PP Removal}
  & PPR & PP Removal & 14123 / 986 / 1013  \TBstrut \\
  \hline\noalign{\hrule height0.8pt} 
\end{tabular}}
\end{center}
\caption{Sample distribution from 9 atomic tasks in the \texttt{Compositional Datasets} partition of StylePTB.}
\label{tab:styleptb}
\end{table*}

\subsection{Dataset}
We first evaluate the functional compositionality of PLMs on the recently released compositional style-transfer dataset, StylePTB\footnote{https://github.com/lvyiwei1/StylePTB}~\cite{lyu2021styleptb} which is built upon Penn TreeBank~\cite{marcinkiewicz1994building}. 
As illustrated in Table~\ref{tab:styleptb}, each atomic task in StylePTB is either a syntactic or semantic style transfer of a single sentence such as changing the tense or removing certain phrases. 
The biggest advantage of StylePTB is that it offers labeled data for many composite tasks from various combinations of atomic tasks. For example, it contains data for TFU (to future tense) and PTA (to active voice), but also TFU+PTA (to future tense in active voice).
For our experiments, we use the \texttt{Compositional Datasets} partition of StylePTB. It consists of all composite tasks and their atomic components, excluding every atomic task that is not composed. As a result, we use 9 \textbf{atomic} tasks and 22 valid \textbf{composite} tasks. The total list of 22 valid composite tasks is found in Table~\ref{tab:prompt_list} or Table~\ref{tab:main_full}.

We also experiment with cross-lingual abstractive summarization on the WikiLingua~\cite{ladhak2020wikilingua}\footnote{https://github.com/esdurmus/Wikilingua}, which gathered multi-lingual guides and their summary from the WikiHow website. The purpose of the experiment with this dataset is to verify whether learned task-composing skill within StylePTB is generalizable to a combination of more realistic and difficult tasks. Out of 10 languages in WikiLingua, we use only two from which the basic T5 can already translate to English: French and German~\cite{raffel2020exploring}\footnote{For StylePTB, the official data split is used. On the other hand, for the WikiLingua dataset, we randomly divide the dataset with an 8:1:1 ratio, using them for train, valid, and test splits respectively because the data splits are not provided publicly for French and German.}.

\subsection{Training Strategies}
\label{subsec:zero_shot_composition}
\begin{table*}[t!]
\begin{center}
\resizebox{0.7\textwidth}{!}{
\begin{tabular}{l l l l}
  \hline\noalign{\hrule height0.8pt} 
  Target: \textcolor{red}{\textbf{A}}+\textcolor{blue}{\textbf{B}} & Strategy & Description & Seen Tasks \TBstrut \\
  \hline \noalign{\hrule height0.8pt} 
  
  \multirow{2}{*}{Zero-Shot}
  & \textsc{Two Atomics}
  & the minimal subset of atomic tasks 
  & \textcolor{red}{\textbf{A}}, \textcolor{blue}{\textbf{B}} \TBstrut \\
  \cline{2-4}
  & \textsc{All Atomics} 
  & all atomic tasks 
  & + [\textbf{C}, \textbf{D}, \textbf{E}, ... ] \TBstrut \\
  \hline \noalign{\hrule height0.8pt} 
  
  \multirow{6}{*}{\shortstack{Zero-Shot \\ (L2C)}} 
  & \multirow{2}{*}{\textsc{Unseen Both}}
  & \multirow{2}{*}{\shortstack[l]{all compositions that does not include \\ any component of the target}}
  & \multirow{2}{*}{+ [\textbf{C}+\textbf{D}, \textbf{C}+\textbf{E}, \textbf{D}+\textbf{E} ... ]}  \TBstrut \\
  & & & \TBstrut \\
  \cline{2-4}
  & \multirow{2}{*}{\textsc{Unseen One} (\textcolor{red}{\textbf{A}})}
  & \multirow{2}{*}{\shortstack[l]{all compositions that does not include \\ one component of the target, \textcolor{red}{\textbf{A}}}}
  & \multirow{2}{*}{+ [\textcolor{blue}{\textbf{B}}+\textbf{C}, \textbf{D}+\textcolor{blue}{\textbf{B}}, ... ]}  \TBstrut \\
  & & & \TBstrut \\
  \cline{2-4}
  & \multirow{2}{*}{\textsc{Hold-1-Out}}
  & \multirow{2}{*}{all compositions other than the target}
  & \multirow{2}{*}{+ [\textbf{E}+\textcolor{red}{\textbf{A}}, \textcolor{red}{\textbf{A}}+\textbf{D}, ... ]} \TBstrut \\
  & & & \TBstrut \\
  \hline \noalign{\hrule height0.8pt} 

  \multirow{1}{*}{Full-Shot} 
  & \textsc{Full}
  & all compositions
  & + [\textcolor{red}{\textbf{A}}+\textcolor{blue}{\textbf{B}}] \TBstrut \\
  \hline \noalign{\hrule height0.8pt} 

\end{tabular}}
\end{center}
\caption{Training strategies regarding data usage with descriptions. There are totally six options, and each row stands for one option. As shown in the last column, the set of seen tasks is accumulated from the top to the bottom. Therefore, the set of training data strictly increases as the row goes down.}
\label{tab:training_strategies}
\end{table*}
One of the most important considerations is that, how many and which atomic/composite tasks are required to learn how to compose arbitrary tasks. In other words, we suggest a systematic way to analyze the type and measure the amount of data needed during the preparatory stage, rather than simply counting sample numbers. Suppose the target composite task is ($A + B$). Here, as illustrated in Table~\ref{tab:training_strategies}, we design 6 training strategies, in increasing order of the number of tasks that are exposed to the model:
\begin{itemize}
    \setlength\itemsep{-0.3em}
    \item \textsc{Two Atomics} shows only the two atomic tasks, $A$ and $B$. It is the harshest setting in our experiments. The model is evaluated on a unique composition of the two atomic tasks.
    \item \textsc{All Atomics} shows all atomic tasks but without any composite tasks. In comparison with \textsc{Two Atomics}, this strategy will highlight the impact of the number of seen atomic tasks. 
    \item \textsc{Unseen Both} provides all atomic tasks and some composite tasks, where composite tasks that share any atomic tasks with the target composition are excluded. 
    \item \textsc{Unseen One ($A$)} is similar to \textsc{Unseen Both}, but only excludes the composite tasks that include the atomic task $A$ of target composition. 
    \item \textsc{Hold-1-Out} includes all composite tasks except only the target composite task. By comparing with \textsc{Unseen Both} and \textsc{Unseen One}, we can check the impact of knowing how the atomic tasks are used in other composite tasks during training. 
    \item \textsc{Full} includes all atomic tasks and all composite tasks.
\end{itemize}
We divide the strategies into three big categories: 1) Zero-Shot, 2) Zero-Shot (L2C), and 3) Full-Shot. Zero-Shot doesn't allow any composite tasks in training while Zero-Shot (L2C) allows some composite tasks except the target composite task. Full-Shot provides the target composite task in training, which can be used as an upper-bound performance.
Each composition method (\textsc{Prompt}, \textsc{Prefix}, and \textsc{Pipeline}) can be trained with the training strategies. However, as mentioned in Section~\ref{sec:method}, \textsc{Prefix} cannot apply Zero-Shot, and \textsc{Pipeline} cannot apply Zero-Shot (L2C) and Full-Shot.

\subsection{Training Details}
For experiments, we follow the hyper-parameters from huggingface T5~\footnote{https://huggingface.co/docs/transformers/model\_doc/t5}. Specifically, we train \texttt{t5-base} with a batch size of 16  for StylePTB dataset. 
We train the model with a learning rate of $5e-5$ using the AdamW optimizer until convergence.
For learning objectives, we cast all the tasks into a ``text-to-text'' format and train them with a maximum likelihood objective: 
\begin{align}
    \max_{\phi} \log Pr_{\phi} (Y | X), 
\end{align}
where $X$ and $Y$ denote the input and output token sequences, and $\phi$ is the set of model parameters.
To avoid catastrophic forgetting of atomic tasks, the training is done in a multi-task manner with a mixed-task batch.
The average time for training is 1 hour. 

For the WikiLingua dataset, we follow the hyperparameter settings from~\cite{chi2021mt6}. 
\texttt{t5-base} is trained with a batch size of 32. The average time for training is 24 hours, and 4 GTX 2080ti's are used. For \textsc{Prefix}, we additionally train approximately 48M parameters. The result is collected via single-run training and evaluation.
\begin{table*}[t!]
\begin{center}
\resizebox{0.85\textwidth}{!}{
\begin{tabular}{l l cccccccc c}
  \hline
  \noalign{\hrule height0.8pt} 
  \multicolumn{1}{l}{\multirow{2}{*}{}} & 
  \multicolumn{1}{l}{\multirow{3}{*}{Model}} &
  \multicolumn{8}{c}{Target Composition (number of samples)} & \multicolumn{1}{c}{\multirow{3}{*}{Avg.}} \TBstrut\\
  \cline{3-10}
  & & \shortstack{PPR+PTA \\ (959)} & \shortstack{TPR+PBF \\ (162)} & \shortstack{TFU+PPR \\ (4492)} & \shortstack{PPR+ATP \\ (1330)} & \shortstack{ARR+PFB \\ (178)} & \shortstack{TFU+PTA \\ (2967)} & \shortstack{TFU+ATP \\ (2455)} & \shortstack{TFU+PFB \\ (233)} & \DTBstrut \\
  \hline\noalign{\hrule height0.8pt} 
  \multirow{2}{*}{\textbf{Full-shot}}
  & \textsc{Prompt} & 96.25 & 93.75 & 89.12 & 84.40 & 64.71 & 88.80 & 83.40 & 82.61 & \databar{87.59} \TBstrut \\ 
  & \textsc{Prefix} & 87.50 & 93.75 & 87.96 & 76.60 & 47.06 & 85.33 & 79.92 & 82.61 & \databar{83.99} \TBstrut \\
  \hline\noalign{\hrule height0.8pt} 
  \multirow{2}{*}{\textbf{Zero-shot}}   
  & \textsc{Pipeline} & 97.50 & 93.75 & 87.50 & 81.56 & 88.24 & 86.87 & 82.63 & 82.61 & \databar{86.55}\TBstrut \\
  & \textsc{Prompt} & 3.75 & 75.00 & 75.93 & 1.42 & 23.53 & 6.95 & 40.54 & 82.61 & \databar{39.31} \TBstrut \\
  \hline\noalign{\hrule height0.8pt}
  \multirow{3}{*}{\shortstack{\textbf{Zero-shot} \\ \textbf{(L2C)}}}
  & \textsc{Prompt} & 95.00 & 93.75 & 89.12 & 12.06 & 70.59 & 86.10 & 83.78 & 86.96 & \databar{79.57}
 \TBstrut \\
 & \textsc{Prompt (GPT-2)} & 50.00 & 87.50 & 55.32 & 33.33 & 11.76 & 57.53 & 43.24 & 69.57 & \databar{50.89}
 \TBstrut \\
  & \textsc{Prefix} & 62.50 & 87.50 & 85.19 & 26.95 & 47.06 & 70.66 & 65.64 & 86.96 & \databar{69.82} \TBstrut \\
  \hline
  \noalign{\hrule height0.8pt} 
\end{tabular}}
\end{center}
\caption{The exact match (EM) scores in percentage on composite tasks from StylePTB. \textbf{Full-shot} models are trained with both all atomic tasks and all composite tasks. \textbf{Zero-shot} models learn all atomic tasks only. \textbf{Zero-shot (L2C)} models learn all atomic tasks and all composite tasks, except the target composite task (\textsc{Hold-1-Out}). Scores are weighted by test sample size of each task to take average. \textbf{Zero-shot (L2C)} models achieve better performance than \textbf{Zero-shot} models, showing the possibility of learning to compose tasks. We evaluate the exact match (EM) scores for each task and take average across tasks using test sample sizes as weights. See appendix Table~\ref{tab:main_full} for the full report including 22 composite tasks. 
}
\label{tab:main}
\end{table*}

\section{Results and Discussion}

We perform intensive experiments to answer five research questions (RQ), where each of them is a title of following subsections.

\subsection{RQ1: Can PLMs compose tasks?}
\label{subsec:lm_can_zeroshot}

We first evaluate whether T5 can compose the already acquired functions on StylePTB dataset, where the results are presented in Table~\ref{tab:main}.
Overall, we empirically confirmed that T5 struggles to compose already acquired functions, where the Zero-shot \textsc{Prompt} fails drastically in some cases, which is consistent with the results in Table~\ref{tab:intro}. 
Though there are some successful cases of showing comparable performance with Full-shot models, it gives only a partial answer to our first research question of asking functional compositionality to language models. 

On the other hand, it is noteworthy that \textsc{Pipeline} shows the second-best score among the methods, which drops only 0.01 points from \textsc{Prompt} of full-shot training on average, even outperforming in some tasks like ``ARR+PFB'' task. 
It demonstrates that \textsc{Pipeline} is the strongest zero-shot baseline as mentioned above. 
However, it is manually composed by humans and the models still cannot know how to compose such tasks. Table~\ref{tab:pipe_order} shows that it is also important for human to carefully choose a proper order between the tasks.

\begin{table*}[t!]
\begin{center}
\resizebox{0.85\textwidth}{!}{ 
\begin{tabular}{l ccccccccc c}
  \hline
  \noalign{\hrule height0.8pt} 
  \multicolumn{1}{l}{\multirow{3}{*}{}} & 
  \multicolumn{8}{c}{Target Composition (number of samples)} &
  \multicolumn{1}{c}{\multirow{3}{*}{Avg.}}
  \TBstrut\\
  \cline{2-9}
  & \shortstack{PPR+PTA \\ (959)} & \shortstack{PPR+ATP \\ (1330)} & \shortstack{TFU+PTA \\ (2967)} & \shortstack{TFU+ATP \\ (2455)} & \shortstack{TPR+PTA \\ (1561)} & \shortstack{TPR+ATP \\ (2163)} & \shortstack{TPA+PTA \\ (1617)} & \shortstack{TPA+ATP \\ (658)} & \DTBstrut \\
  \hline\noalign{\hrule height0.8pt} 
  \textsc{Voice First} & 97.50 & 81.56 & 82.63 & 86.87 & 83.33 & 89.36 & 85.71 & 67.69 & \databar{85.27}

 \TBstrut \\
  \textsc{Voice Later} & 2.50	& 1.42 & 77.99 & 79.54 & 79.63 & 59.04 & 30.00 & 44.62 &  \databar{55.55}
  \TBstrut \\

  
  
  \hline
  \noalign{\hrule height0.8pt} 
\end{tabular}}
\end{center}
\caption{The exact match (EM) scores of \textsc{Pipeline} with different order of computation. 8 target tasks in this table is the set of all compositions that includes a component task from \textit{Voice} category, \textsc{PTA} or \textsc{ATP}. Two annotations \textsc{Voice First} or \textsc{Voice Later} specify the order of components to be applied. For example, \textsc{Voice First} option with a target task PPR+PTA means we perform PTA first, and then do PPR later.}
\label{tab:pipe_order}
\end{table*}

\subsection{RQ2: Can PLMs learn how to compose?}
\label{subsec:lm_can_l2c}
\label{subsec:param_efficient_l2c}

\textbf{Zero-shot (L2C)} results show that a language model \textit{can learn how to compose} tasks, by training some number of compositions and then generalize the mixing mechanism to unseen combinations of atomic tasks. 
Compared to Zero-shot, the \textbf{Zero-shot (L2C)} \textsc{Prompt} performance improves over 100\%, and drops around 10\% compared to Full-shot \textsc{prompt} setting. 
It is noteworthy that the Zero-shot (L2C) setting does not provide any training data for the target task.
We can also see that the same approach considerably well works for GPT2, but not as drastic.

Finally, \textbf{Zero-shot (L2C)} \textsc{prefix} shows that this observation is also valid for such a parameter-efficient model architecture.
However, there is a significant performance drop compared to \textsc{prompt} in general. Another observation in Figure~\ref{fig:adding_comp} is that \textsc{prompt} converges faster than \textsc{prefix}. 
One possible explanation is that learning to compose is difficult enough to require full power of large PLMs.

\subsection{RQ3: Important factors for L2C? }
\label{subsec:better_l2c}

\paragraph{Number of seen composite tasks}

As mentioned in Section~\ref{subsec:lm_can_l2c}, language models can learn how to compose if it is trained with an adequate set of atomic tasks and their combinations.
However, it is infeasible to train all combinations, which is exponentially many, so there comes up with the question on how many is enough.

We provide extra detail for the experiment to evaluate the effect of the number of composite tasks on Zero-shot (L2C) performance. 
We first randomly shuffle the list of 22 composite tasks in StylePTB. 
Cutting until the first $n=0,2,4, \ldots$ elements of the list, we get a sequence of increasing pool of composite tasks, $S_0 \leq S_2 \leq \ldots \leq S_{20}$. 
For each $n$, we basically train the model with $S_n$ and evaluate tasks in $S_n$. 
However, for demonstration, we bound $n$ by $14$ and show evaluation results on the complement set of $S_{14}$, containing 8 tasks, to see the trend. 
\begin{figure}[!t]
  \centering
  \includegraphics[clip,width=0.95\columnwidth]{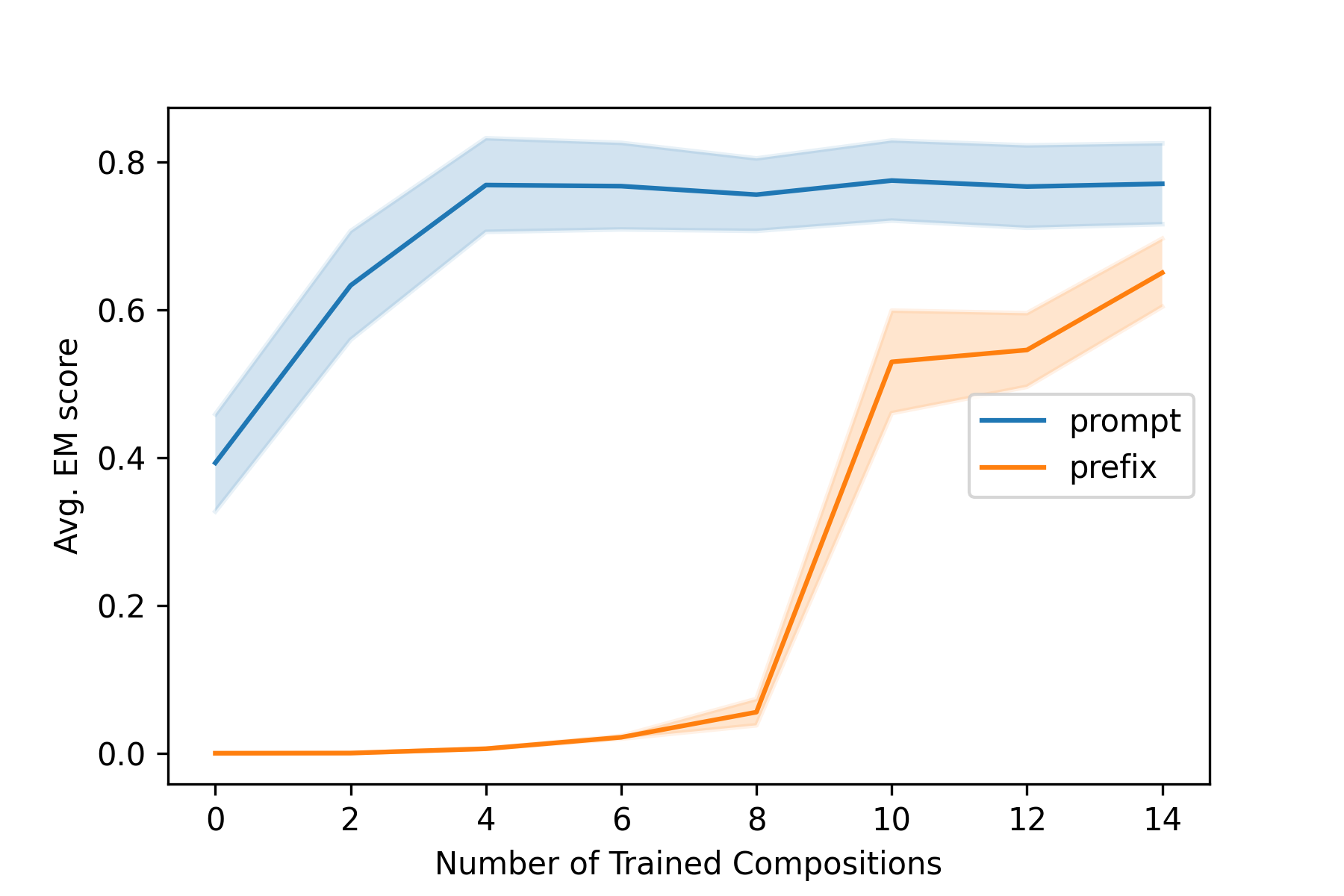}
  \caption{Zero-shot (L2C) average EM scores with respect to number of seen composite tasks. We add two new composite tasks at once and evaluate performance of two models, \textsc{Prompt} and \textsc{Prefix}, on a fixed set of 8 unseen tasks.}
  \label{fig:adding_comp}
\end{figure}
\begin{figure}[!t]
  \centering
  \includegraphics[clip,width=0.95\columnwidth]{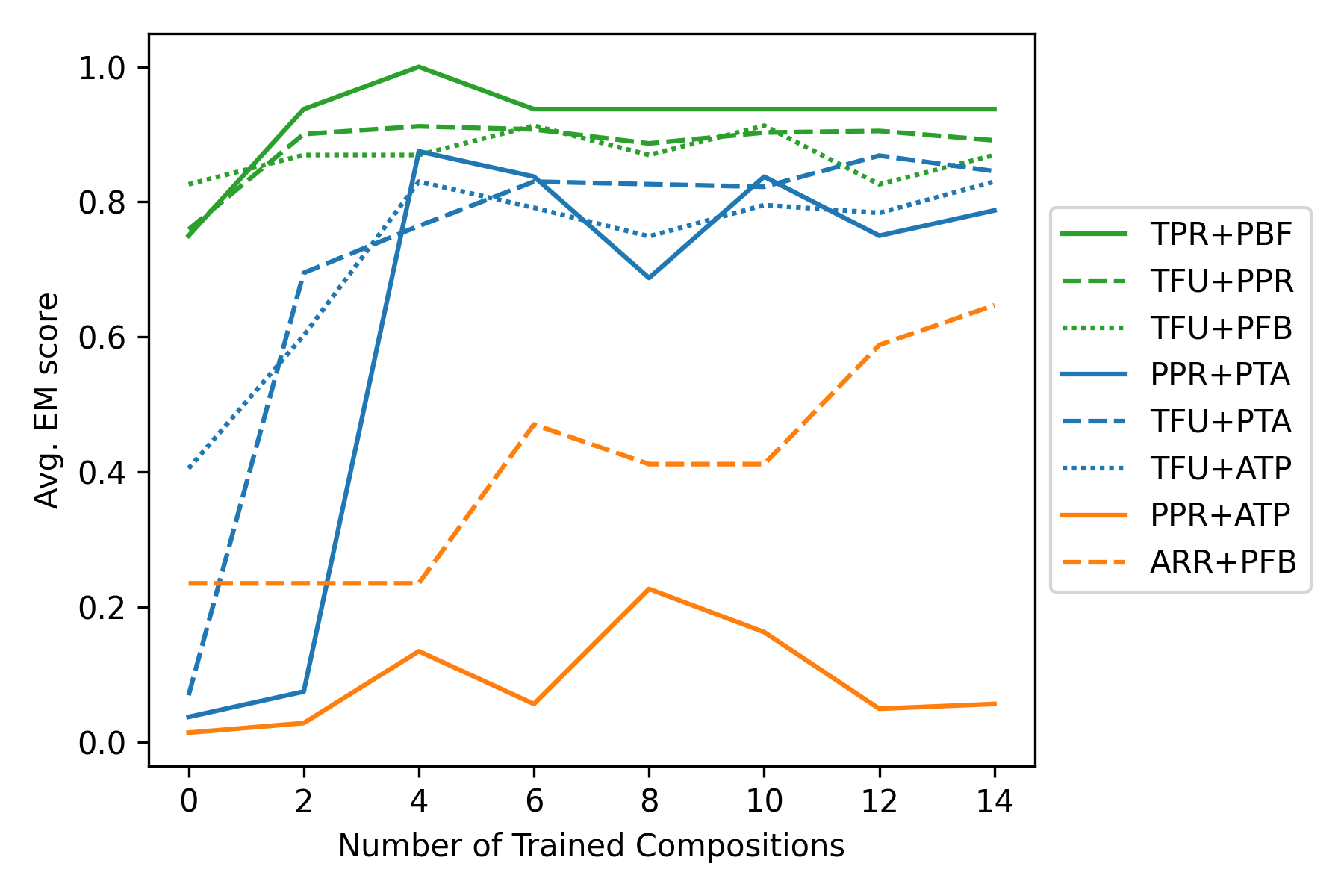}
  \caption{Zero-shot (L2C) exact match (EM) scores on eight target composite tasks with respect to the number of seen composite tasks. The model is \textsc{Prompt}.}
  \label{fig:adding_comp_eight}
\end{figure}

Figures~\ref{fig:adding_comp} and~\ref{fig:adding_comp_eight} indicate that increasing the number of composite tasks for L2C significantly increases the performance as we expected. 
We gradually increase the number of trained compositions from 0 to 14 as described above. 
Figure~\ref{fig:adding_comp_eight} has individual results per task while Figure~\ref{fig:adding_comp} shows averaged results among 8 unseen composite tasks. 

\paragraph{Choice of seen composite tasks}
We observed that more seen composite tasks in training data increase the ability to generalize to unseen composite tasks. 
However, the scenario of adding more tasks totally depends on the permutation of the task sequence. 
Assuming that not only the number of seen composite tasks but also the \textbf{choice} matters, we conduct an ablation study. 
We adopt more logical data restriction strategies described in Section~\ref{subsec:zero_shot_composition}. 
Following the rules, for each target composition out of the 22, an increasing sequence of training datasets is built. 
Then, models are tuned differently depending on those strategies and evaluated on the target task. 
The general effect of each strategy on Zero-shot composition ability is evaluated by averaging out the result through all target tasks.

\begin{table*}[t!]
\begin{center}
\resizebox{0.85\textwidth}{!}{
\begin{tabular}{l cccccccc c}
  \hline
  \noalign{\hrule height0.8pt} 
  \multicolumn{1}{l}{\multirow{3}{*}{Training Strategy}} &
  \multicolumn{8}{c}{Target Composition (number of samples)} & \multicolumn{1}{c}{\multirow{3}{*}{Avg.}} \TBstrut\\
  \cline{2-9}
  & \shortstack{PPR+PTA \\ (959)} & \shortstack{TPR+PBF \\ (162)} & \shortstack{TFU+PPR \\ (4492)} & \shortstack{PPR+ATP \\ (1330)} & \shortstack{ARR+PFB \\ (178)} & \shortstack{TFU+PTA \\ (2967)} & \shortstack{TFU+ATP \\ (2455)} & \shortstack{TFU+PFB \\ (233)} & \DTBstrut \\
  \hline\noalign{\hrule height0.8pt} 
   \textsc{Two Atomics} & 1.25 & 6.25 & 53.94 & 0.71 & 0.00 & 0.00 & 4.25 & 73.91 & \databar{21.37} \TBstrut \\ 
   \textsc{All Atomics} & 3.75 & 75.00 & 75.93 & 1.42 & 23.53 & 6.95 & 40.54 & 82.61 & \databar{39.31}
   \TBstrut \\
   \textsc{Unseen Both} & 42.50 & 93.75 & 85.65 & 21.99 & 17.65 & 78.38 & 72.20 & 82.61 & \databar{70.61} \TBstrut \\
   \textsc{Unseen One (First)} & 78.75 & 87.50 & 90.28 & 1.42 & 0.00 & 83.40 & 85.33 & 82.61 & \databar{76.18} \TBstrut \\
   \textsc{Unseen One (Second)} & 90.00 & 93.75 & 87.73 & 56.03 & 88.24 & 78.38 & 74.90 & 86.96 & \databar{80.03} \TBstrut \\
   \textsc{Hold-1-Out} & 95.00 & 93.75 & 89.12 & 12.06 & 70.59 & 86.10 & 83.78 & 86.96 & \databar{79.57} \TBstrut \\
   \textsc{Full} & 96.25 & 93.75 & 89.12 & 84.40 & 64.71 & 88.80 & 83.40 & 82.61 & \databar{87.59} \TBstrut \\
  
  
  
  \hline\noalign{\hrule height0.8pt} 
\end{tabular}}
\end{center}
\caption{The exact match (EM) scores in percentage, especially focused on comparing training strategies while model is fixed with \textsc{Prompt}. Rows are sorted in strictly increasing order in terms of training data. Average score is weighted by test sample size of each task. For the full results, see appendix Table~\ref{tab:gradual_full}}
\label{tab:gradual}
\end{table*}

\begin{figure}[!t]
  \centering
  \includegraphics[clip,width=\columnwidth]{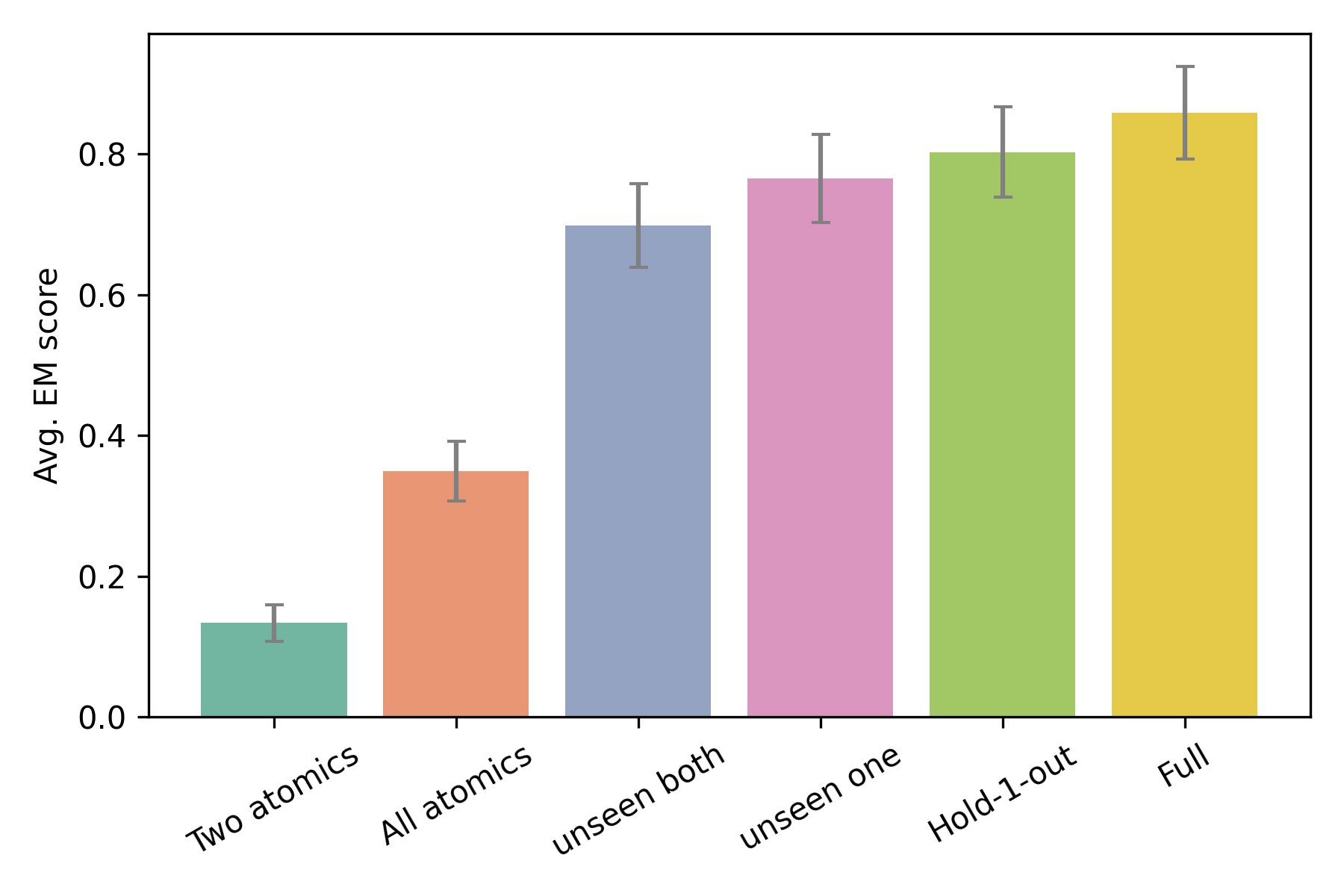}
  \caption{Average EM scores for variants of training strategies with \textsc{Prompt} methods.}
  \label{fig:training_strategies}
\end{figure}

The result is shown in Table~\ref{tab:gradual}. 
Most of the cases, the EM score increases with the level of composite task disclosure. 
As in Figure~\ref{fig:training_strategies}, such monotonicity is clearer in the average EM score. 
Note that the mean score of \textsc{Unseen One (First)} and \textsc{Unseen One (Second)} is still lower than the score of \textsc{Hold-1-out} \footnote{For those tasks where full-shot is worse than zero-shot, dataset errors made during synthetic generation might let additional data not beneficial beyond certain amount.}.

We observe same trend even with the controlled training data size. Table~\ref{tab:gradual_control} shows the result. All training strategies that belong to Zero-shot (L2C) are compared, while a randomly sampled subset of fixed size is used as a training dataset for each option. We can confirm that the EM score still increases as the level of composite task disclosure increases.
\begin{table*}[t!]
\begin{center}
\resizebox{0.85\textwidth}{!}{
\begin{tabular}{l cccccccc c}
  \hline
  \noalign{\hrule height0.8pt} 
  \multicolumn{1}{l}{\multirow{3}{*}{Training Strategy}} &
  \multicolumn{8}{c}{Target Composition (number of samples)} & \multicolumn{1}{c}{\multirow{3}{*}{Avg.}} \TBstrut\\
  \cline{2-9}
  & \shortstack{PPR+PTA \\ (959)} & \shortstack{TPR+PBF \\ (162)} & \shortstack{TFU+PPR \\ (4492)} & \shortstack{PPR+ATP \\ (1330)} & \shortstack{ARR+PFB \\ (178)} & \shortstack{TFU+PTA \\ (2967)} & \shortstack{TFU+ATP \\ (2455)} & \shortstack{TFU+PFB \\ (233)} & \DTBstrut \\
  \hline\noalign{\hrule height0.8pt} 
   \textsc{Unseen Both} & 42.50 & 93.75 & 85.65 & 21.99 & 17.65 & 78.38 & 72.20 & 82.61 & \databar{70.61} \TBstrut \\
   \textsc{Unseen One (First)} & 88.75 & 93.75 & 87.50 & 9.22 & 5.88 & 85.33 & 79.92 & 82.61 & \databar{66.62} \TBstrut \\
   \textsc{Unseen One (Second)} & 90.00 & 93.75 & 87.96 & 48.23 & 70.59 & 82.24 & 77.22 & 86.96 & \databar{80.40} \TBstrut \\
   \textsc{Unseen One (Avg)} & 89.37 & 93.75 & 87.73 & 28.72 & 38.24 & 83.78 & 78.57 & 84.78 & \databar{78.37} \TBstrut \\
   \textsc{Hold-1-Out} & 78.75 & 100.00 & 90.05 & 30.50 & 70.59 & 83.40 & 81.47 & 82.61 & \databar{79.53} \TBstrut \\
  
  
  
  \hline\noalign{\hrule height0.8pt} 
\end{tabular}}
\end{center}
\caption{The exact match (EM) scores in percentage on composite tasks from StylePTB, especially focused on comparing training strategies while the number of training samples is fixed. The model is fixed with \textsc{Prompt}. Rows are sorted in strictly increasing order in terms of training data. Average score is weighted by test sample size of each task.}
\label{tab:gradual_control}
\end{table*}

\subsection{RQ4: Can learned task-composing skills be transferred to other difficult benchmarks?}
One may ask whether the functional compositionality can be transferred to other benchmarks. If the model truly learns how to compose, it can compose any unseen combination of atomic tasks even from different domain. In our setting, this general question is reduced as whether a T5 model that additionally learned Zero-shot (L2C) from StylePTB can compose two pre-trained tasks, summarization and translation.

Table~\ref{tab:intro} shows that for that case Zero-shot (L2C) performance is almost same with Zero-shot. This result indicates that learned task-composing skills is transferable to a limited set of compositions. \ref{subsec:better_l2c} supports this observation more. This limitation motivates a new research direction for large PLMs to achieve human-level generalizability.

\subsection{RQ5: Do larger LMs have more functional compositionality?}
In our preliminary experiments, we observe a very slight chance of GPT-3~\cite{brown2020language} performing functional compositions in a zero-shot manner. 
For example, when we give a manually written prompt ``\texttt{What is the one-sentence French translation of \{text\}? Please answer in one sentence:}'', GPT-3 outputs the French summary of the given text. 
However, such observations require extensive manual prompt tuning.
Furthermore, they cannot generalize to other instances, showing just broken results of performing one of the atomic tasks, yielding an English summary or French translation.
It is thus recommended to further explore the ability of recent extremely large language models, from GPT-3~\cite{brown2020language} to Megatron-Turing~\cite{smith2022using}.

\begin{table}[t!]
\begin{center}
\resizebox{\columnwidth}{!}{
\begin{tabular}{lcccc}
  \hline
  \noalign{\hrule height0.8pt} 
  \multicolumn{1}{l}{\multirow{2}{*}{Model}} &
  \multicolumn{2}{c}{XLS (En-De)} & 
  \multicolumn{2}{c}{XLS (En-Fr)}
  \TBstrut\\
  \cline{2-3} \cline{4-5}
  & ROUGE-4 & ROUGE-L & ROUGE-4 & ROUGE-L \TBstrut\\
  \hline\noalign{\hrule height0.8pt} 
  Fine-tune & 3.14 & 32.63 & 4.45 & 35.56 \TBstrut \\
  \hline
  Pipeline & 3.20 & 32.35 & 3.90 & 33.68 \TBstrut \\
  Zero-shot & 0.43 & 17.05 & 1.10 & 22.32 \TBstrut \\
  Zero-shot (L2C) & 0.43 & 16.98 & 1.13 & 22.43 \TBstrut \\
  \hline
  \noalign{\hrule height0.8pt} 
\end{tabular}}
\end{center}
\caption{Cross-lingual summarization results in English-to-French \& English-to-German WikiLingua XLS~\cite{ladhak2020wikilingua}. We trained \texttt{t5-base}~\cite{raffel2020exploring} on English Summarization and the above translations with prompts in a multi-task learning manner. Note that the ``Zero-shot'' and ``Pipeline'' are trained only with the atomic tasks (translation and summarization), while ``Fine-tune'' model is also further trained with direct cross-lingual summaries. Details about training strategies are listed in the Table~\ref{tab:training_strategies}.
}
\label{tab:intro}
\end{table}
\section{Future Directions}



\paragraph{Pre-training with Pipeline} 
We see great potential for future work utilizing pipeline-based pseudo-labels in the context of functional compositionality. 
Given the positive results we have observed in terms of noisy few-shot training, we are interested in pre-training language models that can learn how to compose seen tasks. 
As recent language models have achieved better and better performances on various single (or, component) tasks, pre-training will benefit from pipeline systems. 

\paragraph{Decomposition in Pre-training}
As studied \cite{lyu2021styleptb}, even a well-defined task can be decomposed into multiple sub-tasks. 
For example, reading comprehension requires recognizing named entities or events in the text, resolving coreferences of them, and selecting an answer among them.
However, recent pre-training strategies, specifically T5, treat it as an atomic task, simply forming an input text as ``\texttt{question: \{question\} context: \{context\}}''. 
In this paper, we argue that giving procedural information of each task in T5-style pre-training, like ``\texttt{entity recognition, coreference resolution, and answer ranking for answering the question: \{question\} context: \{context\}}'', would be helpful to equip language models with functional compositionality and explainability~\cite{kojima2022large}.

\section{Limitations}


Recent extremely large language models, such as GPT-3 and Megatron, are not thoroughly covered in this paper due to limitations in resources. 
For simplicity, we limited our work to compositions of ``pure functions'' meaning that there are no side-effects generated by the functions.
Thus, it is difficult to immediately apply our approach to all NLP pipelines (e.g. Task-oriented Dialogue Systems, classical NLP pipelines, etc.).

Another limitation is that our experiment is limited to ``text-to-text'' models so that it is easier to define compositions as the input and output types are equivalent.
Considering these jointly restricted our scope of work to a certain set of problems.

\section{Conclusion}
This paper explores whether PLMs can compose the functions they have already learned. Our empirical results suggest that 1) PLMS cannot compose as it is, 2) but it can be partially learned (L2C), and 3) the learned task-composing skill is not transferable to other benchmarks, from style transfer to cross-lingual summarization. 
From the results, we suggest several future research directions of pre-training strategies to achieve functional compositionality (\eg, pre-training with pipeline and decomposition in pre-training).

\clearpage
\bibliography{anthology}

\clearpage
\appendix
\section{Appendix}
\label{sec:appendix}

\begin{table*}[t!]
\begin{center}
\resizebox{\textwidth}{!}{
\begin{tabular}{l c l l}
  \hline
  \noalign{\hrule height0.8pt} 
  Dataset & Type & Task & Prompt \TBstrut \\
  \hline\noalign{\hrule height0.8pt} 
  \multirow{31}{*}{\textbf{StylePTB}}
  & \multirow{9}{*}{Atomic}    
  & PPR & \texttt{PPR:} \TBstrut \\
  & & PTA & \texttt{PTA:} \TBstrut \\
  & & ATP & \texttt{ATP:} \TBstrut \\
  & & TFU & \texttt{TFU:} \TBstrut \\
  & & TPR & \texttt{TPR:} \TBstrut \\
  & & TPA & \texttt{TPA:} \TBstrut \\
  & & ARR & \texttt{ARR:} \TBstrut \\
  & & PBF & \texttt{PBF:} \TBstrut \\
  & & PFB & \texttt{PFB:} \TBstrut \\
  \cline{2-4}
  & \multirow{22}{*}{Composition}
  & PPR+ATP & \texttt{PPR + ATP:} \TBstrut \\
  & & PPR+PTA & \texttt{PPR + PTA:} \TBstrut \\
  & & TFU+ATP & \texttt{TFU + ATP:} \TBstrut \\
  & & TFU+PTA & \texttt{TFU + PTA:} \TBstrut \\
  & & TPR+ATP & \texttt{TPR + ATP:} \TBstrut \\
  & & TPR+PTA & \texttt{TPR + PTA:} \TBstrut \\
  & & TPA+ATP & \texttt{TPA + ATP:} \TBstrut \\
  & & TPA+PTA & \texttt{TPA + PTA:} \TBstrut \\
  & & TFU+PPR & \texttt{TFU + PPR:} \TBstrut \\
  & & TPR+PPR & \texttt{TPR + PPR:} \TBstrut \\
  & & TPA+PPR & \texttt{TPA + PPR:} \TBstrut \\
  & & ARR+PFB & \texttt{ARR + PFB:} \TBstrut \\
  & & ARR+PBF & \texttt{ARR + PBF:} \TBstrut \\
  & & TFU+ARR & \texttt{TFU + ARR:} \TBstrut \\
  & & TPA+ARR & \texttt{TPA + ARR:} \TBstrut \\
  & & TPR+ARR & \texttt{TPR + ARR:} \TBstrut \\
  & & TFU+PBF & \texttt{TFU + PBF:} \TBstrut \\
  & & TFU+PFB & \texttt{TFU + PFB:} \TBstrut \\
  & & TPA+PFB & \texttt{TPA + PFB:} \TBstrut \\
  & & TPA+PBF & \texttt{TPA + PBF:} \TBstrut \\
  & & TPR+PBF & \texttt{TPR + PBF:} \TBstrut \\
  & & TPR+PFB & \texttt{TPR + PFB:} \TBstrut \\  
  \cline{2-4}
  \hline\noalign{\hrule height0.8pt} 
  \multirow{5}{*}{\textbf{WikiLingua}}
  & \multirow{3}{*}{Atomic}
  & Summarization & \texttt{summarize:} \TBstrut \\
  & & Translation (en-fr) & \texttt{translate\_en\_fr:}  \TBstrut \\
  & & Translation (en-de) & \texttt{translate\_en\_de:} \TBstrut \\
  \cline{2-4}
  & \multirow{2}{*}{Composition}
  & XLS (en-fr) & \texttt{summarize + translate\_en\_fr:}  \TBstrut \\
  & & XLS (en-de) & \texttt{summarize + translate\_en\_de:} \TBstrut \\
  \hline\noalign{\hrule height0.8pt} 
\end{tabular}}
\end{center}
\caption{List of prompts used to train language models on StylePTB and WikiLingua. Suppose there are two different atomic tasks and corresponding two prompts. Then, the prompt for the new task, defined by their composition, is just a concatenation of their prompts with the `+' symbol in between. Empirically, this template-based prompt composition performed better than many natural writings.}
\label{tab:prompt_list}
\end{table*}
\begin{table*}[t!]
\begin{center}
\resizebox{0.9\textwidth}{!}{
\begin{tabular}{l l cccccccc }
  \hline
  \noalign{\hrule height0.8pt} 
  \multicolumn{1}{l}{\multirow{2}{*}{}} & 
  \multicolumn{1}{l}{\multirow{3}{*}{Model}} &
  \multicolumn{8}{c}{Target Composition (number of samples)} \TBstrut\\
  \cline{3-10}
  & & \shortstack{PPR+PTA \\ (959)} & \shortstack{TPR+PBF \\ (162)} & \shortstack{TFU+PPR \\ (4492)} & \shortstack{PPR+ATP \\ (1330)} & \shortstack{ARR+PFB \\ (178)} & \shortstack{TFU+PTA \\ (2967)} & \shortstack{TFU+ATP \\ (2455)} & \shortstack{TFU+PFB \\ (233)} \DTBstrut \\
  \hline\noalign{\hrule height0.8pt} 
  \multirow{2}{*}{\textbf{Full-shot}}
  & \textsc{Prompt} & 96.25 & 93.75 & 89.12 & 84.40 & 64.71 & 88.80 & 83.40 & 82.61 \TBstrut \\ 
  & \textsc{Prefix} & 87.50 & 93.75 & 87.96 & 76.60 & 47.06 & 85.33 & 79.92 & 82.61 \TBstrut \\
  \hline\noalign{\hrule height0.8pt} 
  \multirow{2}{*}{\textbf{Zero-shot}}   
  & \textsc{Pipeline} & 97.50 & 93.75 & 87.50 & 81.56 & 88.24 & 86.87 & 82.63 & 82.61 \TBstrut \\
  & \textsc{Prompt} & 3.75 & 75.00 & 75.93 & 1.42 & 23.53 & 6.95 & 40.54 & 82.61 \TBstrut \\
  \hline\noalign{\hrule height0.8pt}
  \multirow{3}{*}{\shortstack{\textbf{Zero-shot} \\ \textbf{(L2C)}}}
  & \textsc{Prompt} & 95.00 & 93.75 & 89.12 & 12.06 & 70.59 & 86.10 & 83.78 & 86.96
 \TBstrut \\
 & \textsc{Prompt (GPT-2)} & 50.00 & 87.50 & 55.32 & 33.33 & 11.76 & 57.53 & 43.24 & 69.57 
 \TBstrut \\
  & \textsc{Prefix} & 62.50 & 87.50 & 85.19 & 26.95 & 47.06 & 70.66 & 65.64 & 86.96 \TBstrut \\
  \hline
  \noalign{\hrule height0.8pt} 

  \multicolumn{1}{l}{\multirow{2}{*}{}} & 
  \multicolumn{1}{l}{\multirow{1}{*}{}} &
  \multicolumn{1}{c}{} \TBstrut\\
  & & \shortstack{TPR+ATP \\ (1561)} & \shortstack{TPA+PBF \\ (61)} & \shortstack{ARR+PBF \\ (178)} & \shortstack{TFU+PBF \\ (245)} & \shortstack{TPR+PFB \\ (171)} & \shortstack{TFU+ARR \\ (2166)} & \shortstack{TPR+PTA \\ (2163)} & \shortstack{TPA+ARR \\ (1444)} \DTBstrut \\
  \hline\noalign{\hrule height0.8pt} 
  \multirow{2}{*}{\textbf{Full-shot}}
  & \textsc{Prompt} & 83.33 & 100.00 & 64.71 & 83.33 & 94.12 & 79.04	& 88.30 & 75.00  \TBstrut \\ 
  & \textsc{Prefix} & 80.86 & 100.00 & 76.47 & 91.67 & 76.47 & 64.19 & 85.11 & 72.85  \TBstrut \\
  \hline\noalign{\hrule height0.8pt} 
  \multirow{2}{*}{\textbf{Zero-shot}}   
  & \textsc{Pipeline} & 83.33 & 100.00 & 94.12 & 87.50 & 88.24 & 77.73 & 89.36 & 78.81 \TBstrut \\
  & \textsc{Prompt} & 32.10 & 	66.67 & 17.65 & 50.00 & 94.12 & 3.93 & 10.64 & 0.0464  \TBstrut \\
  \hline\noalign{\hrule height0.8pt}
  \multirow{3}{*}{\shortstack{\textbf{Zero-shot} \\ \textbf{(L2C)}}}
  & \textsc{Prompt} & 84.57 & 100.00 & 76.47 & 83.33 & 88.24 & 75.11 & 79.26 & 81.46
 \TBstrut \\
 & \textsc{Prompt (GPT-2)} & 45.68 & 83.33 & 47.06 & 83.33 & 58.82 & 32.31 & 63.83 & 67.55
 \TBstrut \\
  & \textsc{Prefix} & 74.07 & 100.00 & 47.06 & 87.50 & 88.24 & 61.57 & 67.02 & 0.6556
 \TBstrut \\
  \hline
  \noalign{\hrule height0.8pt} 

  \multicolumn{1}{l}{\multirow{2}{*}{}} & 
  \multicolumn{1}{l}{} &
  \multicolumn{1}{c}{} \TBstrut\\
  & & \shortstack{TPA+PFB \\ (70)} & \shortstack{TPA+PTA \\ (1617)} & \shortstack{TPA+PPR \\ (658)} & \shortstack{TPA+PPR \\ (1926)} & \shortstack{TPR+PPR \\ (3054)} & \shortstack{TPR+ARR \\ (1260)} & \shortstack{Avg \\ (29350)} & \DTBstrut \\
  \hline\noalign{\hrule height0.8pt} 
  \multirow{2}{*}{\textbf{Full-shot}}
  & \textsc{Prompt} & 100.00	& 93.57 & 76.92 & 91.35 & 87.33 & 75.00 & \databar{85.85} & \TBstrut \\ 
  & \textsc{Prefix} & 100.00 & 87.14 & 69.23 & 87.57 & 82.88 & 63.64 & \databar{81.03} & \TBstrut \\
  \hline\noalign{\hrule height0.8pt} 
  \multirow{2}{*}{\textbf{Zero-shot}}   
  & \textsc{Pipeline} & 100.00 & 85.71 & 67.69 & 89.73 & 85.27 & 77.27 & \databar{84.88} & \TBstrut \\
  & \textsc{Prompt} & 100.00 & 5.71 & 7.69 & 76.22 & 73.97 & 3.79 & \databar{34.92} & \TBstrut \\
  \hline\noalign{\hrule height0.8pt}
  \multirow{3}{*}{\shortstack{\textbf{Zero-shot} \\ \textbf{(L2C)}}}
  & \textsc{Prompt} & 100.00 & 82.86 & 46.15 & 90.81 & 86.30 & 71.97 & \databar{80.27} &
 \TBstrut \\
 & \textsc{Prompt (GPT-2)} & 71.43 & 55.00 & 23.08 & 70.27 & 66.10 & 52.27 & \databar{53.94} &
 \TBstrut \\
  & \textsc{Prefix} & 100.00 & 64.29 & 30.77 & 83.78 & 81.51 & 65.15 & \databar{70.02} & \TBstrut \\
  \hline
  \noalign{\hrule height0.8pt} 
\end{tabular}}
\end{center}
\caption{The exact match (EM) scores in percentage on composite tasks from StylePTB. \textbf{Full-shot} models are trained with both all atomic tasks and all composite tasks. \textbf{Zero-shot} models learn all atomic tasks only. \textbf{Zero-shot (L2C)} models learn all atomic tasks and all composite tasks, except the target composite task. Scores are weighted by test sample size of each task to take average. We evaluate the exact match (EM) scores for each task and take average across tasks using test sample sizes as weights. 
}
\label{tab:main_full}
\end{table*}

\begin{table*}[t!]
\begin{center}
\resizebox{0.9\textwidth}{!}{
\begin{tabular}{l cccccccc}
  \hline
  \noalign{\hrule height0.8pt} 
  \multicolumn{1}{l}{\multirow{3}{*}{Training Strategy}} &
  \multicolumn{8}{c}{Target Composition (number of samples)} \TBstrut\\
  \cline{2-9}
  & \shortstack{PPR+PTA \\ (959)} & \shortstack{TPR+PBF \\ (162)} & \shortstack{TFU+PPR \\ (4492)} & \shortstack{PPR+ATP \\ (1330)} & \shortstack{ARR+PFB \\ (178)} & \shortstack{TFU+PTA \\ (2967)} & \shortstack{TFU+ATP \\ (2455)} & \shortstack{TFU+PFB \\ (233)} \DTBstrut \\
  \hline\noalign{\hrule height0.8pt} 
   \textsc{Two Atomics} & 1.25 & 6.25 & 53.94 & 0.71 & 0.00 & 0.00 & 4.25 & 73.91 \TBstrut \\ 
   \textsc{All Atomics} & 3.75 & 75.00 & 75.93 & 1.42 & 23.53 & 6.95 & 40.54 & 82.61
   \TBstrut \\
   \textsc{Unseen Both} & 42.50 & 93.75 & 85.65 & 21.99 & 17.65 & 78.38 & 72.20 & 82.61 \TBstrut \\
   \textsc{Unseen One (First)} & 78.75 & 87.50 & 90.28 & 1.42 & 0.00 & 83.40 & 85.33 & 82.61 \TBstrut \\
   \textsc{Unseen One (Second)} & 90.00 & 93.75 & 87.73 & 56.03 & 88.24 & 78.38 & 74.90 & 86.96 \TBstrut \\
   \textsc{Hold-1-Out} & 95.00 & 93.75 & 89.12 & 12.06 & 70.59 & 86.10 & 83.78 & 86.96 \TBstrut \\
   \textsc{Full} & 96.25 & 93.75 & 89.12 & 84.40 & 64.71 & 88.80 & 83.40 & 82.61 \TBstrut \\
  
  
  
  \hline\noalign{\hrule height0.8pt} 

  \multicolumn{1}{l}{\multirow{3}{*}{Training Strategy}} &
  \multicolumn{8}{c}{Target Composition (number of samples)} \TBstrut\\
  \cline{2-9}
  & \shortstack{TPR+ATP \\ (1561)} & \shortstack{TPA+PBF \\ (61)} & \shortstack{ARR+PBF \\ (178)} & \shortstack{TFU+PBF \\ (245)} & \shortstack{TPR+PFB \\ (171)} & \shortstack{TFU+ARR \\ (2166)} & \shortstack{TPR+PTA \\ (2163)} & \shortstack{PTA+ARR \\ (1444)} \DTBstrut \\
  \hline\noalign{\hrule height0.8pt} 
   \textsc{Two Atomics} & 29.01 & 0.00 & 11.76 & 58.33 & 88.24 & 0.44 & 0.00 & 0.66 \TBstrut \\ 
   \textsc{All Atomics} & 32.10 & 66.67 & 17.65 & 50.00 & 94.12 & 3.93 & 10.64 & 4.64
   \TBstrut \\
   \textsc{Unseen Both} & 79.01 & 100.00 & 17.65 & 83.33 & 100.00 & 44.54 & 70.74 & 54.97 \TBstrut \\
   \textsc{Unseen One (First)} & 83.95 & 100.00 & 11.76 & 87.50 & 94.12 & 75.55 & 74.47 & 68.21 \TBstrut \\ 
   \textsc{Unseen One (Second)} & 78.40 & 100.00 & 52.94 & 79.17 & 94.12 & 58.52 & 72.87 & 58.94 \TBstrut \\
   \textsc{Hold-1-Out} & 84.57 & 100.00 & 76.47 & 83.33 & 88.24 & 75.11 & 79.26 & 81.46 \TBstrut \\
   \textsc{Full} & 83.33 & 100.00 & 64.71 & 83.33 & 94.12 & 79.04 & 88.30 & 75.00 \TBstrut \\
  
  
  
  \hline\noalign{\hrule height0.8pt} 

  \multicolumn{1}{l}{\multirow{3}{*}{Training Strategy}} &
  \multicolumn{6}{c}{Target Composition (number of samples)} & \multicolumn{1}{c}{\multirow{3}{*}{Avg.}} & \TBstrut\\
  \cline{2-7}
  & \shortstack{TPA+PFB \\ (70)} & \shortstack{TPA+PTA \\ (1617)} & \shortstack{TPA+ATP \\ (658)} & \shortstack{TPA+PPR \\ (1926)} & \shortstack{TPR+PPR \\ (3054)} & \shortstack{TPR+ARR \\ (1260)} & &  \DTBstrut \\
  \hline\noalign{\hrule height0.8pt} 
   \textsc{Two Atomics} & 71.43 & 0.00 & 1.54 & 9.19 & 24.66 & 0.76 & \databar{15.39} & \TBstrut \\ 
   \textsc{All Atomics} & 3.75 & 75.00 & 75.93 & 1.42 & 23.53 & 6.95 & \databar{34.92} & \TBstrut \\
   \textsc{Unseen Both} & 100.00 & 73.57 & 41.54 & 86.30 & 84.59 & 55.30 & \databar{69.80} & \TBstrut \\
   \textsc{Unseen One (First)} & 100.00 & 84.29 & 50.77 & 91.89 & 87.33 & 75.00 & \databar{77.96} & \TBstrut \\
   \textsc{Unseen One (Second)} & 100.00 & 75.00 & 41.54 & 88.65 & 83.56 & 43.94 & \databar{75.06} & \TBstrut \\
   \textsc{Hold-1-Out} & 100.00 & 82.86 & 46.15 & 90.81 & 86.30 & 71.97 & \databar{80.28} &  \TBstrut \\
   \textsc{Full} & 100.00 & 93.57 & 76.92 & 91.35 & 87.33 & 75.00 & \databar{85.85} & \TBstrut \\
  
  
  
  \hline\noalign{\hrule height0.8pt} 
\end{tabular}}
\end{center}
\caption{The exact match (EM) scores in percentage on composite tasks from StylePTB, especially focused on comparing training strategies while model is fixed with \textsc{Prompt}. The results for all composite tasks are in Appendix Figure~\ref{fig:training_strategies}. Rows are sorted in strictly increasing order in terms of training data. Average score is weighted by test sample size of each task.}
\label{tab:gradual_full}
\end{table*}

\end{document}